\pdfoutput=1

\documentclass[11pt]{article}

\usepackage[final]{acl}

\usepackage{times}
\usepackage{latexsym}

\usepackage[T1]{fontenc}

\usepackage[utf8]{inputenc}

\usepackage{microtype}

\usepackage{inconsolata}

\usepackage{graphicx}

\usepackage{times}
\usepackage{latexsym}
\usepackage{bm}
\usepackage{hyperref}
\usepackage{color,colortbl}
\usepackage{framed}
\definecolor{shadecolor}{rgb}{0.92,0.92,0.92}
\usepackage{stfloats}
\usepackage{enumitem}
\usepackage{booktabs}
\usepackage{nccmath}
\usepackage{multirow}
\usepackage{subfigure}
\usepackage{todonotes}
\usepackage{makecell}
\usepackage{dashrule}
\usepackage{amssymb}
\usepackage{bbding}
\usepackage{arydshln}
\usepackage{algorithm}
\usepackage{algorithmic}

\definecolor{green}{rgb}{0.84, 0.91, 0.83}
\definecolor{blue}{rgb}{0.83, 0.88, 0.96}
\definecolor{gr}{rgb}{0.86, 0.86, 0.86}

\usepackage{soul} 

\title{Multi-perspective Alignment for Increasing Naturalness in \\ Neural Machine Translation}

 \author{
  Huiyuan Lai$^1$,
  Esther Ploeger$^2$,
  Rik van Noord$^1$,
  Antonio Toral$^3$\thanks{Work partly carried out while affiliated with the University of Groningen.}\\
  $^1$CLCG, University of Groningen, The Netherlands \\ 
  $^2$Department of Computer Science, Aalborg University, Denmark \\ 
  $^3$Departament de Llenguatges i Sistemes Informàtics, Universitat d'Alacant, Spain \\
  \texttt{\{h.lai,r.i.k.van.noord\}@rug.nl} \hspace{0.8em}
  \texttt{espl@cs.aau.dk} 
  \hspace{0.8em}
  \texttt{antonio.toralr@ua.es}
  }

\begin{document}
\maketitle
\begin{abstract}
Neural machine translation (NMT) systems amplify lexical biases present in their training data, leading to artificially impoverished language in output translations. 
These language-level characteristics render automatic translations different from text originally written in a language and human translations, which hinders their usefulness in for example creating evaluation datasets.
Attempts to increase naturalness in NMT can fall short in terms of content preservation, where increased lexical diversity comes at the cost of translation accuracy.
Inspired by the reinforcement learning from human feedback framework,
we introduce a novel method that rewards both naturalness and content preservation.
We experiment with multiple perspectives to produce more natural translations, aiming at reducing machine and human translationese.
We evaluate our method on English-to-Dutch literary translation, and find that our best model produces translations that are lexically richer and exhibit more properties of human-written language, without loss in translation accuracy.
\end{abstract}

\section{Introduction}
While machine translation (MT) has achieved promising performance with the adoption of neural network~\citep{bahdanau-etal-2015-neural, vaswani-etal-2017-attention, nllb2022},
automatic translations remain markedly different from translations by professional human translators.
A striking example is the fact that MT outputs exhibit reduced lexical diversity~\citep{vanmassenhove-etal-2019-lost,vanmassenhove-etal-2021-machine} and increased source-language interference~\citep{toral-2019-post} compared to human translation (HT).
These linguistic differences were previously referred to as \textit{machine translationese}~\citep{declercq:hal-02454668,bizzoni-etal-2020-human,vanmassenhove-etal-2021-machine}.\footnote{This term has since been criticized, see for example \citet{crespo2023translationese}.}

Within the context of natural language processing (NLP), these language-level artifacts of machine translation can have negative implications.
For example, machine translationese in NLP evaluation datasets can inflate performance assessments. Examples of this are found in MT  \citep{zhang-toral-2019-effect, graham-etal-2020-statistical} and cross-lingual transfer learning~\citep{yu-etal-2022-translate, artetxe-etal-2020-translation}. 
Furthermore, in the field of literary translation, preserving reading experience (and thus the original style) can be an important aspect of the translation process~\citep{delabastita2011literary,toral-way-2015-machine, guerberof2020impact}.

\begin{figure}
    \centering
\includegraphics[scale=.85]{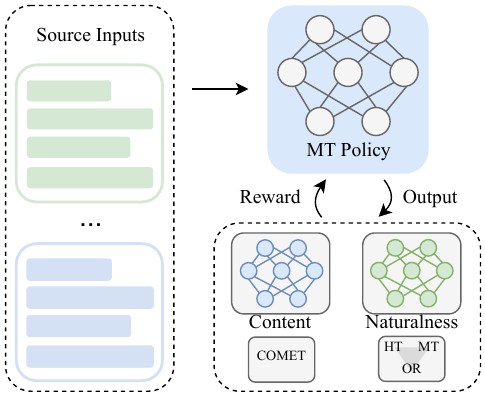}
    \caption{Aligning the translation policy from both content preservation and naturalness perspectives.}
    \label{fig:framework-naturalness}
\end{figure}

Reducing translation artifacts in MT output is not trivial.
Intuitively, translated texts should match the style of the texts originally written in that target language, while preserving the content of the source language.
This trade-off between naturalness and content preservation presents methodological challenges.
For example, previous work shows a decrease in translation quality when aiming to recover lexical diversity in MT~\citep{ploeger-etal-2024-towards}.
Moreover, existing approaches such as Tagging~\citep{freitag-etal-2022-natural}, aim to increase MT naturalness in a rigid manner, while the amount of naturalness in the output translation cannot be manually adjusted to a desired level. Yet, in cases where faithfulness to the source is crucial, the naturalness of a translation may be of lesser importance \citep{parthasarathi-etal-2021-sometimes-want}.

To address these challenges, we frame the task of increasing naturalness in MT as a text style transfer-like task, where style and content are the two core aspects~\citep{mou-vechtomova-2020-stylized, lai-etal-2021-thank, lai-etal-2021-generic}.
In practice, we train a vanilla MT model with supervised learning and subsequently
exploit reward learning that fosters naturalness and content preservation, as shown in Figure~\ref{fig:framework-naturalness}. 
With respect to naturalness we explore two objectives: making MT more akin to human translations (i.e. reducing machine translationese) and making MT more akin to texts originally written in the target language, i.e. reducing translationese~\citep{gellerstam-1986-translationese,baker-1993-corpus,toury2012descriptive}.
We evaluate our framework on a dataset for English-to-Dutch literary translation. Our main contributions are as follows:
\begin{itemize}
    \item We introduce a novel flexible multi-perspective alignment framework that favours natural translation outputs while fostering content preservation;
    \item We experiment with and analyse the results of three different preference classifiers that are used to produce more natural translations: preferring original target-language text (OR) over HT, OR over MT, and HT over MT;
    \item Extensive experiments show that our model produces translations that are lexically richer 
    than baseline MT systems without loss in translation quality.\footnote{All code at \url{https://github.com/laihuiyuan/alignment4naturalness}}
\end{itemize}

\section{Related Work}
\label{sec:related_work}

\subsection{Increasing MT Naturalness}
A few approaches have been put forward to make MT outputs more natural.
For example, \citet{freitag-etal-2019-ape} trained a post-processor that learns to translate from round-trip machine translated text to original text in the same language, which can be applied to the outputs of existing MT systems.
\citet{freitag-etal-2022-natural} prepend their training examples with special tags that denote whether the target side of the training data was originally written in that language or not. 
These methods are rigid, while in some cases, content preservation may be more important than style~\citep{parthasarathi-etal-2021-sometimes-want}. In response, \citet{ploeger-etal-2024-towards} propose a flexible approach based on reranking translation candidates, but report considerable loss in general translation quality. 

In parallel efforts, some research aims to reduce translationese from human translations, and uses monolingual approaches based on text style transfer~\citep{jalota-etal-2023-translating}, semantic parsing~\citep{wein-schneider-2024-lost} and debiasing embeddings~\citep{dutta-chowdhury-etal-2022-towards}. Additionally, there is growing interest in leveraging human feedback to improve overall translation quality where a single metric such as COMET trained from human annotations is used as the reward model~\citep{ramos-etal-2024-aligning, he-etal-2024-improving}. In this work we focus on improving translation quality from multiple perspectives, which is tailorable to the downstream scenario, while still being faithful to the source texts.

\subsection{(Machine) Translation Detection}

Following neural machine translation (NMT), a new line of research started to investigate the extent to which translations (including HT and MT) contain artefacts, and how these compare to original texts and human translations.

\paragraph{HT vs OR Classification}
\citet{baroni2006} showed that original texts can be distinguished from human-translated texts with computational methods.
Concrete textual markers, such as the frequency of function words or the use of punctuation, have been associated with this difference \citep{koppel-ordan-2011-translationese,volansky2015features}.
Beyond hand-crafting specific linguistic features, \citet{pylypenko-etal-2021-comparing} find that neural architectures provide a reliable tool for distinguishing translated from original texts. They obtain state-of-the-art performance by fine-tuning multilingual BERT  \citep{devlin-etal-2019-bert} on the task. 

\paragraph{MT vs HT Classification} 
\citet{bizzoni-etal-2020-human} show that there is a difference between the translation artifacts produced by humans and MT models.
\citet{van-der-werff-etal-2022-automatic} use neural language models to distinguish between HT and NMT in German-to-English translation, and highlight the challenges of this task, with their sentence-level system achieving an accuracy of approximately 65\%.
This is further investigated in a multilingual setting~\citep{chichirau-etal-2023-automatic}.

These works show that HT, MT and original texts can, to some extent, be distinguished from each other
with neural methods. Based on this, we expect that our reward functions with neural classifiers can be effective for improving naturalness in MT outputs.

\section{Data}
\label{sec:data}
In this section, we describe datasets used for (machine) translation detection and MT, including both a parallel and a monolingual corpus of books.
Table~\ref{tab:data_stats} shows the sizes and splits of both datasets.

\begin{table}[t]
\begin{center}
\setlength{\tabcolsep}{4pt}
\resizebox{\columnwidth}{!}{
\begin{tabular}{llrr}
\toprule
\bf Data Split &   \bf Language  & \bf \# Books & \bf \# Sentences\\
\midrule
\multicolumn{4}{c}{\textbf{Translationese Detection}} \\
\midrule
 Train & Dutch (OR)     & 143 & 982,114 \\
       & Dutch (HT) & 143 & 1,390,351\\
\hline
 Test  & Dutch (OR) & 36  & 261,151 \\
       & Dutch (HT)  & 36  &  340,950 \\
\midrule
\multicolumn{4}{c}{\textbf{Machine Translation}} \\
\midrule
 Train & Dutch (HT) & 495 & 4,874,784\\
       & English (OR)    & 495 & 4,874,784\\
\hline
 Valid & Dutch (HT) & 5   & 88,881\\
       & English (OR)   & 5   & 88,881\\
\hline
 Test  & Dutch (HT) & 31  & 302,976\\
       & English (OR)   & 31  & 302,976\\
\midrule
Baseline (Train)& Dutch (OR) & - & 4,874,784\\
Baseline (Valid)& Dutch (OR) & - & 88,881\\
\bottomrule
\end{tabular}}
\caption{Data set division and size.}
\label{tab:data_stats}
\end{center}
\end{table}

\paragraph{Translationese Detection Data}
We use a dataset consisting of books written in Dutch~\citep{toral-etal-2021-literary} from a range of authors and genres, as preprocessed by~\citet{ploeger-etal-2024-towards}.
The dataset contains 7,000 books that were manually annotated to be originally written in Dutch (OR) or in another language (HT).
From these, we derive two balanced subsets: 286 books for training and 72 for testing.

\paragraph{Machine Translation Data}
We use the parallel dataset from~\citet{toral-etal-2021-literary}, preprocessed by~\citet{ploeger-etal-2024-towards}. 
This dataset consists of 531 books that were originally written in English (OR) and human translated into Dutch (HT), of which 495 books for training, 5 for validation and 31 as a test set. The genres of these books vary, 
including
literary fiction, popular fiction, non-fiction and children's books from over 100 authors. Particularly, the test set also contains a broad range of books.\footnote{A full list of author names, titles, genres and publishing years of the test set books can be found in Appendix \ref{app:test}.}
In addition, we use monolingual data for the two baseline MT systems (see Section~\ref{sec:baselines}), consisting of a random sample of equal size to the parallel training data and disjoint from the subset used for translation detection.

\section{Methodology}
In this section, we first introduce the base MT model (Section~\ref{sec:basemt}) and binary translationese classification models (Section~\ref{sec:translat_classif}) using supervised learning.
Subsequently, we propose a multi-perspective alignment framework based on reward learning, which explicitly optimises the MT model with human expectations, aiming to increase naturalness and to preserve content (Section~\ref{sec:mtreduction}).

\subsection{Base MT Model}\label{sec:basemt}
As the initial step of model alignment, we train the base MT model with supervised learning on high-quality parallel data. Specifically, given a source text $x=\{x_{1}, \cdots, x_{n}\}$ of length $n$ in language $l_{s}$ and a target text $y=\{y_{1}, \cdots, y_{m}\}$ of length $m$ in language $l_{t}$ from dataset $\mathcal{D}$, the MT model aims to learn two conditional distributions, transforming $x$ to $y$. We begin with Transformer-based models whose goal is to minimize the following negative log-likelihood:
\begin{equation}
\label{eq:loss-ce}
    \mathcal{L}_{nl} = - \frac{1}{m}{\textstyle \sum_{i=1}^{m} \mathrm{log}\left (p(y_i|y_{0:i-1}, x;\theta) \right)} 
\end{equation}

\noindent Where $\theta$ 
represents
model parameters
and
$y_i$ 
the $i$-th token of the target sequence. 

\subsection{(Machine) Translationese Classification}\label{sec:translat_classif}
We use three different 
classifiers, seeing the promotion of natural translations from
different perspectives, namely preferring OR over HT, HT over MT, and OR over MT. 
The first classifier aims at reducing human translationese, while the second and third ones aim at reducing machine translationese (the second one with respect to HT and the third one with respect to OR).
These classifiers will be used as rewards (Section~\ref{sec:mtreduction}) to foster naturalness. 
Having three perspectives will allow us to find out how each of them impacts the accuracy and naturalness of the resulting translations.

For HT vs OR classification, we use the monolingual Dutch data introduced in the first part of Table~\ref{tab:data_stats}.
For the other two settings, we translated a subset of the English text in the parallel data (second part of Table~\ref{tab:data_stats}) of equal size to the monolingual training data (982,114 sentences)
into Dutch using the base MT model.
The resulting machine translated sentences are combined with OR texts in the monolingual data for MT vs OR classification and with HT texts in the parallel data for MT vs HT.
We filter out machine translated texts that are identical to human translations. Based on the above data, we fine-tune the Dutch language model BERTje~\citep{devries-etal-2019-bertje} for the binary detection tasks, obtaining three different models. 

\begin{algorithm}[tb]
  \caption{Multi-perspective alignment algorithm for Naturalness and Content}\label{alg:rl}
  \small
  \begin{algorithmic}[1]
     \REQUIRE{Base MT model $p(y | x; \theta_0)$, Training set: source $\bm{X}$ and target $\bm{Y}$}
     \REQUIRE{Reward function: COMET $C(x, y, \hat{y})$ and translationese classification $p(t_1|\hat{y};\phi)$}
     \FOR{each step $i=0, 1, \cdots, m$}
       \STATE {$M_i \gets \text{MiniBatch}(\bm{X}, \bm{Y})$}
       \FOR{$x \in M_i$}
         \STATE {$\hat{y} \sim p(y|x;\theta_i)$}
         \STATE {Calc. translationese reward $r_t({\hat{y}})$ by Eq.~\ref{eq:reward-tran}}
         \STATE {Calc. content reward $r_c({\hat{y}})$ by Eq.~\ref{eq:reward-cont}}
         \STATE {Calc. overall reward $r({\hat{y}})$ by Eq.~\ref{eq:reward-overall}}
       \ENDFOR
       \STATE {Update MT model using data $M_i$ and $\hat{M}_i$ with the overall reward based on Eq.~\ref{eq:loss-overall}}
     \ENDFOR
  \end{algorithmic}
\label{al:rl}
\end{algorithm}

\subsection{Multi-perspective Alignment for Naturalness and Content Preservation}
\label{sec:mtreduction}

We introduce our method which 
ranks samples based on rewards that target naturalness and content preservation.
This approach is inspired by recent work in text style transfer, where both meaning has to be preserved and style should be transferred~\citep{lai-etal-2021-thank, lai-etal-2021-generic}. This content vs form trade-off is similar to our situation with content preservation and naturalness. Specifically, 
after training
a base MT model using supervised learning (Section~\ref{sec:basemt}), 
we further align it
with human expectations in terms of naturalness and content in the form of reward learning.

Based on the base MT model, we train our reward learning based framework. The MT model takes source text $x$ as input and generates the corresponding translated text $\hat{y}$. To ensure the quality of the $\hat{y}$, we design two rewards that 
aim to foster 
naturalness and
content preservation. 
We consider the two quality feedbacks as rewards and fine-tune the MT model through reinforcement learning. The overview of our alignment framework is shown in Algorithm~\ref{al:rl}.

\paragraph{Rewarding Naturalness}
We use a binary translationese classifier (OR vs HT, HT vs MT or OR vs MT) to assess how well the translated text $\hat{y}$ scores on the translationese aspect, i.e., to assess its (machine) translationese probability. Formally, this reward is formulated as
\begin{equation}
\label{eq:reward-tran}
   r_{t}(\hat{y}) = 
\begin{cases} 
0 & \text{if } p(t_1|\hat{y};\phi ) < \sigma_{t}   \\
p(t_1|\hat{y};\phi ) & \text{otherwise}
\end{cases} 
\end{equation}

\noindent where $\phi$ is the parameter of the classifier. $\sigma_{t}$ is the translationese threshold, which is 
set to 0.5 in our experiments based on preliminary results.

\paragraph{Rewarding Content} 
We employ COMET~\citep{rei-etal-2020-comet} as the content-based reward model $C(x, y, \hat{y})$ to assess the content quality of $\hat{y}$ as the translation of $x$. This is formulated as
\begin{equation}
\label{eq:reward-cont}
   r_{c}(\hat{y}) = 
\begin{cases} 
0 & \text{if } \mathrm{C}(x, y, \hat{y}) < \sigma_{c}   \\
\mathrm{C}(x, y, \hat{y}) & \text{otherwise}
\end{cases} 
\end{equation}

\noindent Where C($\cdot$) represents the COMET model and $\sigma_{t}$ represents the content threshold, which is set to 0.85 in our experiments based on preliminary results.


\paragraph{Overall Reward}
To encourage the model to 
foster naturalness
while preserving the content, the ﬁnal reward is the harmonic mean of the above two rewards
\begin{equation}
\label{eq:reward-overall}
   r(\hat{y}) = 
\begin{cases} 
0 & \text{if } r_t=0 \text{ or } r_c=0 \\
\frac{2}{1/r_t+1/r_c}  & \text{otherwise}
\end{cases} 
\end{equation}

\paragraph{Learning Objectives} Here we aim to maximize the expected reward of the generated sequence $\hat{y}$, the loss is defined as
\begin{equation}
\label{eq:loss-reward}
    \mathcal{L}_{rw} = - \frac{1}{m}{\sum_{i=1}^{m} r(\hat{y}) \mathrm{log}\left (p(\hat{y}_i|\hat{y}_{0:i-1}, x;\theta) \right)} 
\end{equation}

\noindent To keep the fine-tuned model from moving too far from the base MT model, we combine the reward objective with the supervised training loss instead of using a reference model requiring large computing resources. Therefore, the final objective function of our framework consists of two components: negative log-likelihood loss in Eq.~\ref{eq:loss-ce} and reward-based loss in Eq.~\ref{eq:loss-reward}, which are jointly formulated as 
\begin{equation}
\label{eq:loss-overall}
    \mathcal{L}(\theta; \mathcal{D}) = \mathbb{E}_{(x,y)\sim \mathcal{D}}[\beta \mathcal{L}_{nl} + \mathcal{L}_{rw}]
\end{equation}

\noindent Where $\beta$ a is a hyperparameter used to control the weight of the negative log-likelihood loss 
(set to 0.5 in our main experiments), allowing our method to be tailorable. We employ the policy gradient algorithm~\citep{Williams-1992} to maximize the expected reward.

\section{Experimental Setup}

\subsection{Baselines}\label{sec:baselines}

In addition to the base MT model (Section~\ref{sec:basemt}), we include three previous methods that aim at reducing machine translationese as baselines:
Tailored RR (Top-k)~\citep{ploeger-etal-2024-towards}, 
automatic post-editing (APE)~\citep{freitag-etal-2019-ape} and Tagging~\citep{freitag-etal-2022-natural}. 

\paragraph{Tailored RR} is an approach that involves reranking translation candidates with a classifier that distinguishes between original and translated text. We select the Tailored RR (Top-$k$), which reranks candidates that are obtained through Top-$k$ sampling, as a baseline, since it retrieves the highest diversity in~\citet{ploeger-etal-2024-towards}.

\paragraph{APE} aims to train a post-processor that transforms machine-translated Dutch into more natural Dutch texts. To obtain parallel data of source synthetic Dutch and original Dutch, we round-trip translate the original Dutch text of the monolingual data.

\paragraph{Tagging} aims to learn to differentiate between original and translated texts. We use the base Dutch-English MT model to obtain English translations
of the monolingual original Dutch text. Then, we prepend a tag \texttt{<orig>} to the English text in the above data, \texttt{<tran>} to the English text in the parallel data, and train a new MT model 
on the concatenation of these two datasets.

We include two settings for the amount of original target data (i.e.~\texttt{<orig>}): one equivalent to the parallel training data (4.8M) and the other to the translationese classifier data (1M). This is done to investigate how the proportions of target-translated vs target-original in the training data affect results. Our hypothesis is that the larger the percentage of target-original the more natural the translations, but at the expense of lower translation accuracy.

\subsection{Implementation Details}
All experiments are implemented using the library HuggingFace Transformers~\citep{wolf-etal-2020-transformers}. 
We use 
the
BART~\citep{lewis-etal-2020-bart} architecture with 6 Transformer-based~\citep{vaswani-etal-2017-attention} layers in both the encoder and decoder. 
The base MT models are trained using the AdamW optimiser~\citep{loshchilov2018decoupled} with a cosine learning rate decay, and a linear warmup of 1,000 steps. 
The maximum learning rate is set to 1e-4, the batch size is 256, and the gradient accumulation is 2; all reward-based models are trained with a consistent learning rate of 2e-5.
We evaluate the model every 1,000 steps and use early stopping with patience 3 if the cross-entropy loss on the validation set does not decrease. 

We use beam search with size 5 during 
inference. Since some of the training data contains instances of repeated punctuation marks, this led to the reinforcement learning method tending to optimize the model for higher rewards. Therefore, we take a simple post-processing step to remove consecutive repeated punctuation marks after the text is generated.\footnote{See Appendix~\ref{app:post} for post-processing examples.}

\subsection{Evaluation Methods}
We perform a comprehensive evaluation on the model outputs, including translation quality and translationese evaluation. 
Unless stated otherwise, the scores are reported by taking the averages for all books in the test set.

\paragraph{Translation Quality}
We employ three metrics to automatically calculate the content preservation of the output based on human references (and source sentences), namely BLEU~\citep{papineni-etal-2002-bleu}, COMET~\citep{rei-etal-2020-comet, rei-etal-2022-cometkiwi}, and MetricX~\citep{juraska-etal-2024-metricx}. We use the Sacre-BLEU implementation~\citep{post-2018-call} for BLEU. Regarding the COMET family models, we use both the default model \texttt{wmt22-comet-da} (COMET), and the reference-free model \texttt{wmt22-cometkiwi-da} (KIWI) that is not used for reward learning. For MetricX, we use \texttt{MetricX-24-Hybrid-XL}, considering it our most important translation quality metric, since it achieved state-of-the-art performance at the WMT24 Metrics Shared Task~\citep{freitag-etal-2024-llms}.

\begin{figure*}[!t]
\centering
\includegraphics[scale=.52]{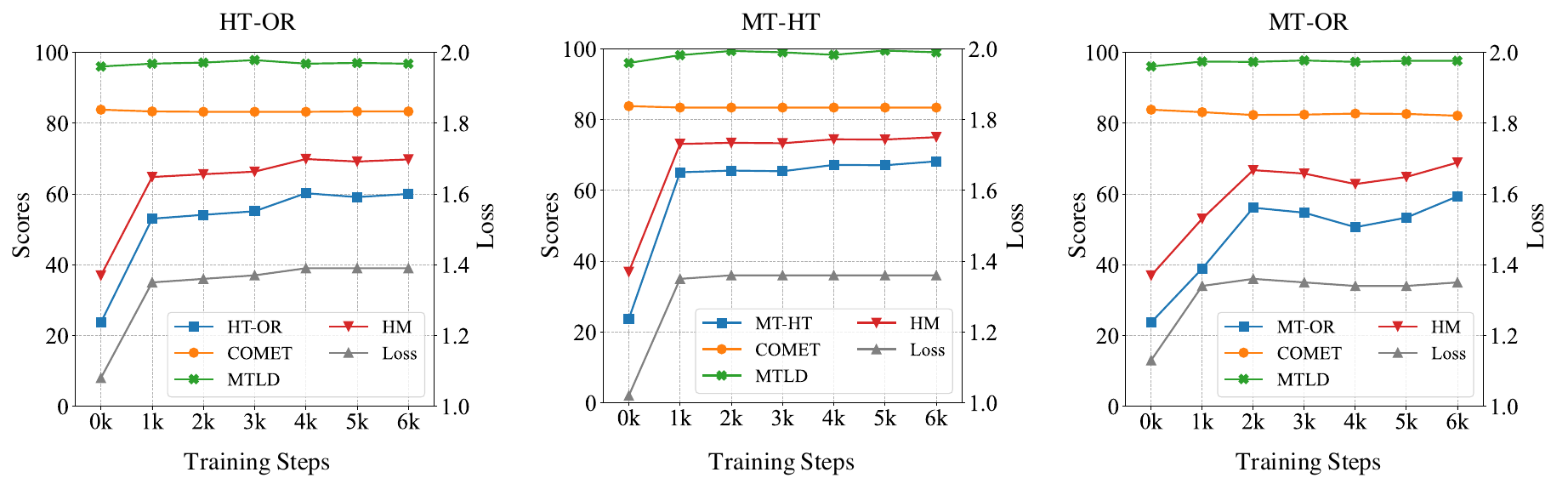}
\caption{Evaluation results on the validation set under various settings. Notes: (i) The training step of 0K represents the 
base
MT model; HM indicates the harmonic mean of classification accuracy and COMET score.} 
\label{fig:training-iteration}
\end{figure*}

\paragraph{Translationese Evaluation}
We apply the translationese detection models to MT outputs and report the rate (i.e.\ classification accuracy) at which they are classified as the target aspect, such as OR in HT-OR, with higher rates indicating that the outputs are more human-like. 
Additionally, as previous studies show that translated texts are often simpler than original texts~\citep{baker-1993-corpus}, 
our evaluation also covers lexical diversity.
Here we report six different metrics:

\begin{itemize}[leftmargin=*]
\itemsep 0in
\item TTR~\citep{templin1957certain}: Type-Token Ratio is the number of unique words (types) divided by the total number of words in the text.

\item Yule's I~\citep{yule1944}: Given the size of the vocabulary (number of types) $V$ and $f(i,N)$ representing the numbers of types which occur $i$ times in a sample of length $N$, Yule's I is calculated as
\begin{equation}
\label{eq:yule}
\text{Yule's I} = \frac{V^2}{{\textstyle \sum_{i=1}^{V}i^2 * f(i,N)}-V} 
\end{equation}

\item MTLD~\citep{philip2005}: evaluated sequentially as the average length of sequential word strings in a text that maintains a given TTR value. We use a threshold of 0.72, following~\citet{vanmassenhove-etal-2021-machine}.
This metric has been shown to be stable across different text lengths~\citep{mccarthy2010mtld}, which is why we consider it more important a metric than TTR or Yule's I.

\item B1~\citep{vanmassenhove-etal-2021-machine}: the percentage of words in the output that are in the estimated 1,000 most frequent words in a language.

\item PTF~\citep{vanmassenhove-etal-2021-machine}: the average percentage (over all relevant source words) of times the most frequent translation option was chosen among all translation options.

\item CDU~\citep{vanmassenhove-etal-2021-machine}: the cosine similarity between the output vector for each source word and a vector of the same length with an equal distribution for each translation option.\footnote{See~\citet{ploeger-etal-2024-towards} for details on its implementation.}

\end{itemize}

\begin{figure}[!t]
\centering
\includegraphics[scale=.42]{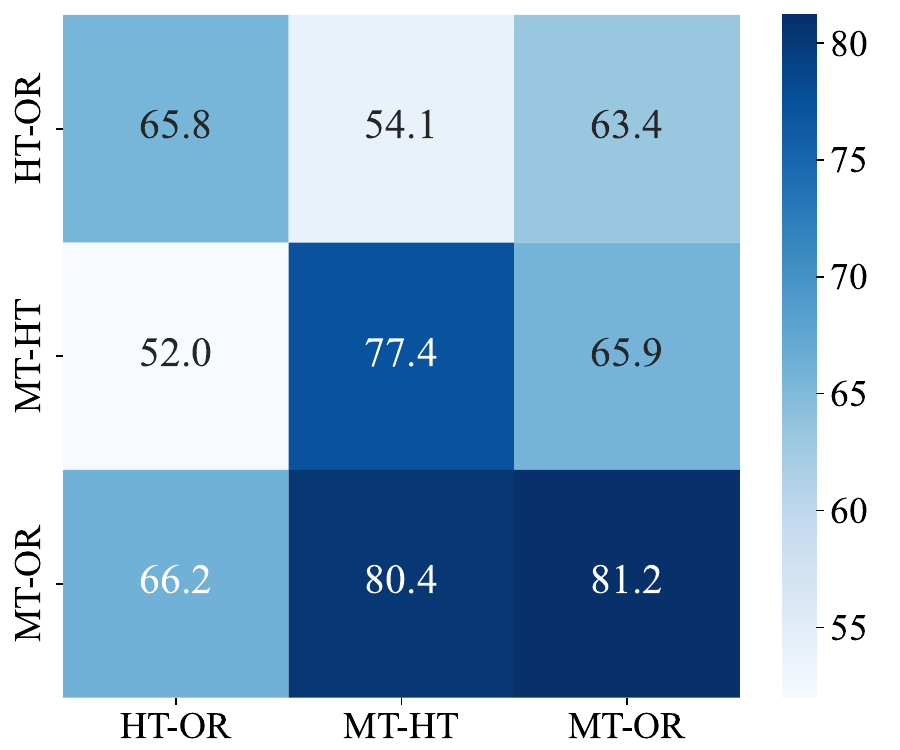}
\caption{Confusion matrices for the different binary classifiers. Each row represents the results of a classifier tested on different test sets.} 
\label{fig:matrix}
\end{figure}

\section{Results and Analysis}

\subsection{Initial Results}

\paragraph{Translationese Classification} Figure~\ref{fig:matrix} 
shows the performance confusion matrix of different binary translationese classifiers on different test sets. We can observe that for each classifier scores on the main diagonal are higher than others, with MT-OR having the highest score, followed by MT-HT and HT-OR. This is on par with the performance in similar scenarios from previous work~\citep{pylypenko-etal-2021-comparing}. While \citet{van-der-werff-etal-2022-automatic} show that the distinction between the translation variants (MT and HT) is challenging, we found that human translations are even more difficult to distinguish from the original target-language texts.
Interestingly, MT-OR achieves higher accuracy on the sets of HT-OR and MT-HT than their corresponding classifiers.

\paragraph{Machine Translation} 
During the alignment phase (see Section~\ref{sec:mtreduction}) we ﬁnd that for some models the valid loss does not correlate with the naturalness aspect (i.e.\ classifier's accuracy) after 1k steps: while naturalness improves, the loss on the validation set stays flat. Therefore we manually select checkpoints between the 1k and the 6k steps, and report their evaluation and loss curves in Figure~\ref{fig:training-iteration}. 

The first observation is that all models achieve substantial improvement in naturalness over the first 1k steps compared to the base MT model (i.e.\ 0K), as reflected in the results for translationese classification (HT-OR, MT-HT and MT-OR) and lexical richness (MTLD). Although the COMET scores of some models decrease slightly, the overall score HM follows the trend of the translationese aspect. After 1k steps, MTLD scores and valid loss tend to be flat; translated language classification shows a clear improvement from 1k to 2k steps on MT-OR, a slight increase on HT-OR, and remains stable on MT-HT. For all models, although some metrics fluctuate after 2k steps, they tend to be stable overall.
For the remaining experiments, we report the results of the alignment model at 5K trainingsteps.

\begin{table*}[!t]
\centering
\setlength{\tabcolsep}{3pt}
\resizebox{\linewidth}{!}{%
\begin{tabular}{lccccccccccccc}
\toprule
 & \multicolumn{4}{c}{\bf Translation Accuracy} & \multicolumn{3}{c}{\bf Classification Accuracy} & \multicolumn{6}{c}{\bf Lexical Diversity}\\
 \cmidrule(lr){2-5}\cmidrule(lr){6-8}\cmidrule(lr){9-14}
\textbf{} & \textbf{BLEU} & \textbf{COMET} & \textbf{KIWI} & \textbf{MetricX$\downarrow$} & \textbf{HT-OR} & \textbf{MT-HT} & \textbf{MT-OR} & \textbf{TTR} & \textbf{Yule's I} & \textbf{MTLD} & \textbf{B1$\downarrow$} & \textbf{PTF$\downarrow$} & \textbf{CDU$\downarrow$}\\
\midrule
Human Translation    & -    & -    & - & - & 32.9 & 69.3 & 48.6 & 0.153 & 3.934 & 96.0 & 0.672 & 0.817 & 0.548\\
\midrule
Tailored RR & 21.2 & 74.5 & 72.4 & 4.86 & \underline{\bf{35.1}} & \underline{\bf{52.9}} & 33.5 & 0.157 & 4.170 & \underline{\bf{104.3}} & 0.682 & \underline{\bf{0.815}} & 0.559\\
APE            & 29.9 & 80.4 & 77.9 & 3.38 & 33.7 & 33.6 & 35.2 & 0.155 & 3.670 & 91.7 & 0.682 & 0.824 & 0.561\\
Tagging (1M)  & 31.6 & 81.6 & 80.1 & 2.87 & 33.0 & 42.6 & 36.9 & 0.161 & 4.133 & 95.8 & 0.671 & 0.817 & \bf{0.554}\\
Tagging (4.8M)  & 31.1 & 80.9 & 79.7 & 3.05 & 33.5 & 43.2 & \underline{\bf{39.0}} & \underline{\bf{0.164}} & \underline{\bf{4.347}} & 96.8 &  \underline{\bf{0.667}} & \underline{\bf{0.815}} & 0.556\\
BM: Base MT Model    & \underline{\bf{32.5}} & \underline{\bf{82.3}} & \bf{80.4} & \bf{2.66} & 28.1 & 18.9 & 17.6 & 0.150 & 3.537 & 90.4 & 0.677 & 0.826 & 0.563\\
\midrule
BM + COMET \& HT-OR  & 29.7 & 80.4 & 79.9 & 2.83 & \bf{34.0} & 24.0 & 25.5 & 0.145 & 3.239 & 91.0 & 0.675 & 0.830 & 0.554\\
BM + COMET \& MT-HT  & \bf{32.1} & \bf{82.2} & \underline{\bf{80.6}} & \underline{\bf{2.63}} & 26.1 & \bf{33.4} & 26.6 & \bf{0.150} & \bf{3.572} & \bf{93.3} & \bf{0.674} & 0.828 & 0.553\\
BM + COMET \& MT-OR  & 31.5 & 81.5 & 80.1 & 2.75 & 28.7 & 33.3 & \bf{28.2} & \bf{0.150} & 3.544 & 91.8 & 0.678 & \bf{0.827} & \underline{\bf{0.542}}\\
\bottomrule
\end{tabular}}
\caption{\label{tab:main}
Translation performance under various settings. Note that bold numbers indicate the best system for each block, and underlined numbers indicate the best score by an MT system for each metric.}
\end{table*}

\subsection{Main Results}\label{sec:mainresults}

We report the main results in Table~\ref{tab:main}, including 
the base MT model, the three baselines and our methods trained with both rewards: COMET for content preservation and the three different classifiers for naturalness (i.e.\ HT-OR, MT-HT and MT-OR).

Tailored RR achieves the highest naturalness scores (e.g.\ HT-OR and MTLD), but performs the worst on all translation accuracy metrics.
Compared to APE, Tagging consistently performs better across the board, both in terms of content (i.e.\ translation accuracy) and naturalness.  Additionally, we observe that using more target-original data (i.e. 4.8M vs 1M) results in lower accuracy scores but better naturalness metrics, which is consistent with our hypothesis (see Section~\ref{sec:baselines}). Overall, we observe that the three baselines underperform the base MT model in terms of translation accuracy and outperform it in most cases when it comes to naturalness metrics.

Moving to our approach, when comparing different classification rewards, the model trained with COMET \& MT-HT achieves, overall, better scores than our other two models (HT-OR and MT-OR). 
We speculate that the rewards that foster OR do not work as well due to a mismatch between the preference of the classifier (OR) and the data in the target side of the MT training data (HT). We thus believe that such classifiers could be useful in scenarios in which the target side of the MT training data contains texts originally written in that language, which would be common in translation directions in which the target language is higher-resourced than the source language.

Overall, our best system (BM + COMET \& MT-HT) achieves better naturalness scores than the base MT model (e.g.\ 93.3 vs 90.4 for MTLD), while even having a higher KIWI score (80.6 vs 80.4) and a lower MetricX score (2.63 vs 2.66; lower is better), two metrics that have not been used in our reward learning. Tagging attains
higher naturalness scores but this comes at the price of a notable reduction in translation accuracy, as shown by KIWI (79.7 vs 80.6) and MetricX (3.05 vs 2.63).

\begin{table*}[!t]
\centering
\setlength{\tabcolsep}{3pt}
\resizebox{\linewidth}{!}{%
\begin{tabular}{lccccccccccccc}
\toprule
 & \multicolumn{4}{c}{\bf Translation Accuracy} & \multicolumn{3}{c}{\bf Classification Accuracy} & \multicolumn{6}{c}{\bf Lexical Diversity}\\
 \cmidrule(lr){2-5}\cmidrule(lr){6-8}\cmidrule(lr){9-14}
\textbf{} & \textbf{BLEU} & \textbf{COMET} & \textbf{KIWI} & \textbf{MetricX$\downarrow$} & \textbf{HT-OR} & \textbf{MT-HT} & \textbf{MT-OR} & \textbf{TTR} & \textbf{Yule's I} & \textbf{MTLD} & \textbf{B1$\downarrow$} & \textbf{PTF$\downarrow$} & \textbf{CDU$\downarrow$}\\
\midrule
BM: Base MT Model   & 32.5 & 82.3 & 80.4 & 2.66 & 28.1 & 18.9 & 17.6 & 0.150 & 3.537 & 90.4 & 0.677 & 0.826 & 0.563\\
BM + COMET          & 32.2 & 81.9 & 80.7 & 2.64 & 26.7 & 19.1 & 19.6 & 0.147 & 3.362 & 90.9 & 0.679 & 0.830 & 0.543\\
\midrule
BM + HT-OR          & 31.1 & 81.0 & 80.0 & 2.75 & 30.3 & 21.5 & 22.1 & 0.137 & 1.950 & 26.8 & 0.700 & 0.826 & 0.556\\
BM + HT-OR \& COMET & 29.7 & 80.4 & 79.9 & 2.83 & 34.0 & 24.0 & 25.5 & 0.145 & 3.239 & 91.0 & 0.675 & 0.830 & 0.554\\
\midrule
BM + MT-HT          & 32.2 & 81.5 & 80.2 & 2.67 & 28.2 & 24.7 & 22.4 & 0.149 & 3.465 & 91.2 & 0.679 & 0.826 & 0.556\\
BM + MT-HT \& COMET & 32.1 & 82.2 & 80.6 & 2.63 & 26.1 & 33.4 & 26.6 & 0.150 & 3.572 & 93.3 & 0.674 & 0.828 & 0.553\\
\midrule
BM + MT-OR          & 32.6 & 81.9 & 80.3 & 2.65 & 26.8 & 22.9 & 22.4 & 0.149 & 3.460 & 90.8 & 0.680 & 0.826 & 0.559\\
BM + MT-OR \& COMET & 31.5 & 81.5 & 80.1 & 2.75 & 28.7 & 33.3 & 28.2 & 0.150 & 3.544 & 91.8 & 0.678 & 0.827 & 0.542\\
\bottomrule
\end{tabular}}
\caption{\label{tab:ablation}
Ablation study: The contribution of each reward component,  where we fine-tune the base MT model using only the content reward or the naturalness reward.
}
\end{table*}

\subsection{Ablation Study}
To assess the contribution of each reward component in our framework, we perform a set of ablation studies, the results of which are shown in Table~\ref{tab:ablation}. For the COMET vs COMET + classifier setting, we see higher naturalness scores in the latter in all cases for MT-HT and MT-OR (except CDU in MT-HT), as expected, while there are mixed cases in HT-OR.
Also as expected, translation accuracy scores decrease when the naturalness reward is added (except COMET with MT-HT).

Compared to models using classifier-only reward, classifier + COMET results generally in better naturalness-related metrics (except PTF), but worse content-based metrics (except COMET with MT-HT). 
This might be due to a mismatch between the classifier's objective and the training data (see reasoning in Section~\ref{sec:mainresults}) and to complex interactions between both rewards, that would require further inspection.

\begin{table*}[!t]
\centering
\setlength{\tabcolsep}{3pt}
\resizebox{\linewidth}{!}{%
\begin{tabular}{lp{15cm}}
\toprule
\textbf{Source} & \textbf{Text}\\
\midrule
Original English              & It was because of the atmosphere of hockey-fields and cold baths and \colorbox{green}{community hikes} and \colorbox{blue}{general clean-mindedness} which she managed to carry about with her.\\
Human Translation   & Het was om de sfeer van hockeyvelden en koude douches en \colorbox{green}{groepsuitstapjes} en \colorbox{blue}{algemene geestelijke reinheid} die zij om zich wist te verspreiden.
\\
\midrule
Tagging (4.8M) & Het kwam door de sfeer van hockeyvelden en koude baden en \colorbox{green}{gemeenschapsfietsen} en \colorbox{blue}{algemeene properheid}, die zij  met haar wist rond te voeren. \\
BM: Base MT Model & Het kwam door de sfeer van hockeyvelden, koude baden en \colorbox{green}{plattelandskantoren} en \colorbox{blue}{algehele schoonheid} die ze met zich mee kon nemen. \\
\midrule
BM + COMET \& MT-HT & Dat kwam door de atmosfeer van hockeyvelden, koude baden en \colorbox{green}{gemeenschapshikes} en \colorbox{blue}{algemene properheid} die ze  met zich mee kon dragen.\\
\bottomrule
\end{tabular}}
\caption{\label{tab:case}
Example of human-written text (source and human translation), translations of the most relevant baselines (Tagging, base MT model) and our alignment model (BM + COMET \& MT-HT).
}
\end{table*}

\begin{table*}[!t]
\centering
\setlength{\tabcolsep}{3pt}
\resizebox{\linewidth}{!}{%
\begin{tabular}{lccccccccccccc}
\toprule
 & \multicolumn{4}{c}{\bf Translation Accuracy} & \multicolumn{3}{c}{\bf Classification Accuracy} & \multicolumn{6}{c}{\bf Lexical Diversity}\\
 \cmidrule(lr){2-5}\cmidrule(lr){6-8}\cmidrule(lr){9-14}
\textbf{} & \textbf{BLEU} & \textbf{COMET} & \textbf{KIWI} & \textbf{MetricX$\downarrow$} & \textbf{HT-OR} & \textbf{MT-HT} & \textbf{MT-OR} & \textbf{TTR} & \textbf{Yule's I} & \textbf{MTLD} & \textbf{B1$\downarrow$} & \textbf{PTF$\downarrow$} & \textbf{CDU$\downarrow$}\\
\midrule
Human Translation    & -    & -    & -   & - & 32.9 & 69.3 & 48.6 & 0.153 & 3.934 & 96.0 & 0.672 & 0.817 & 0.548\\
BM: Base MT Model    & 32.5 & 82.3 & 80.4 & 2.66 & 28.1 & 18.9 & 17.6 & 0.150 & 3.537 & 90.4 & 0.677 & 0.826 & 0.563\\
\midrule
BM + COMET \& HT-OR  & 21.8 & 78.0 & 77.5 & 3.59 & 43.5 & 48.4 & 42.8 & 0.138 & 2.859 & 88.0 & 0.674 & 0.848 & 0.527\\
BM + COMET \& MT-HT  & 24.1 & 81.3 & 79.6 & 3.06 & 27.0 & 52.2 & 34.6 & 0.121 & 2.265 & 92.4 & 0.683 & 0.849 & 0.547\\
BM + COMET \& MT-OR  & 24.4 & 80.5 & 79.8 & 3.19 & 32.2 & 59.2 & 49.5 & 0.139 & 3.084 & 93.1 & 0.669 & 0.845 & 0.526\\
\bottomrule
\end{tabular}}
\caption{\label{tab:hyper}
Translation performance with $\beta$ 
 set to 0.0, where models are trained without the constraint of negative log-likelihood loss.}
\end{table*}

\subsection{Finer-grained Analysis} 

\paragraph{Surface-level Inspection}
\label{sec:surface} 
In Table~\ref{tab:case}, we compare the surface-level output of the strongest baseline (Tagging; 4.8M) with that of the base MT model and our best alignment system (COMET \& MT-HT).
As highlighted in \colorbox{green}{green}, the English `community hikes' is translated to \textit{gemeenschapsfietsen} (`community bicycles') by the Tagging system, while our alignment system outputs \textit{gemeenschapshikes} (`community \textit{hikes}'). 
This is an example of how the Tagging model output may score high on lexical diversity metrics, but strays from the content, where our model preserves it.
As shown in \colorbox{blue}{blue}, `general clean-mindedness' is translated to \textit{algehele schoonheid} (`overall beauty') by the base MT system. Our alignment system translates to \textit{algemene properheid} (`general cleanliness'), while the Tagging system outputs \textit{algemeene properheid}. The latter case contains a double \textit{e}, which is not typical in this context for modern Dutch, but does appear in the original Dutch dataset. Our alignment MT system is not affected by this. 

\begin{figure*}[!t]
\centering
\includegraphics[scale=.57]{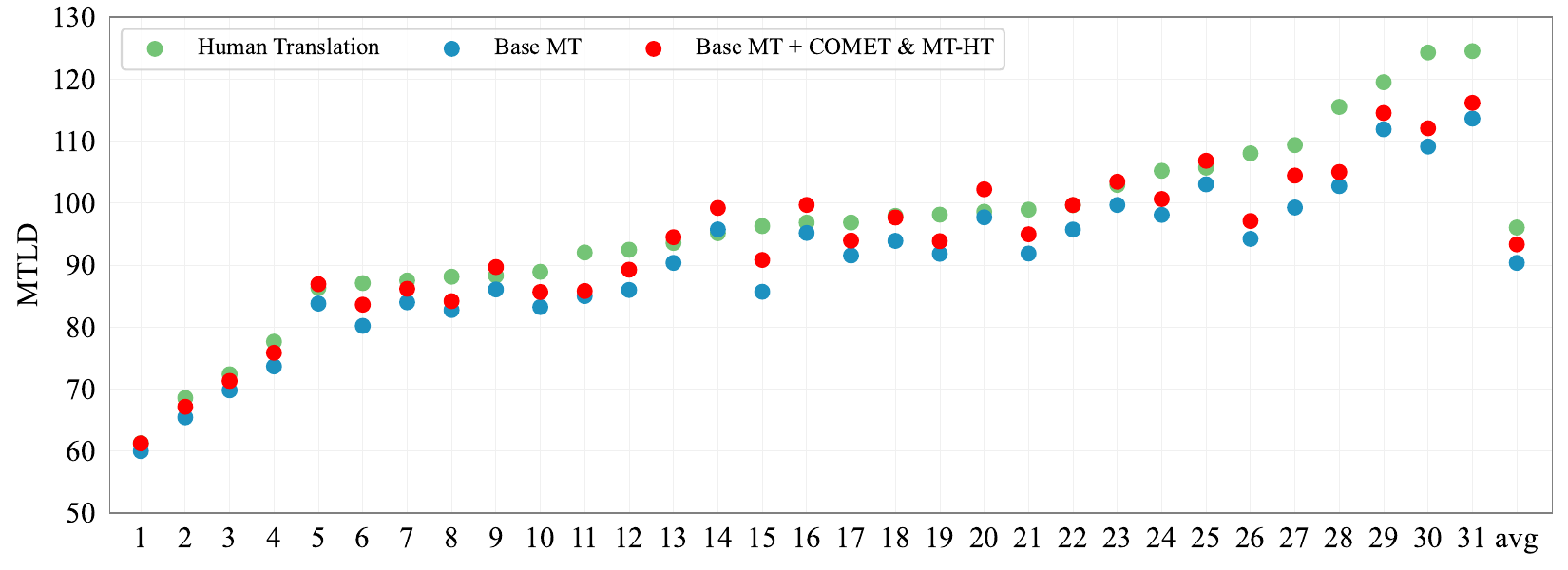}
\caption{Per-book comparison of MTLD under human translation, base MT model, and alignment model. Note that avg presents the average score across all books.}
\label{fig:mtld}
\end{figure*}

\paragraph{Book-level Comparison} 
Figure~\ref{fig:mtld} shows MTLD scores per book between human translation, base MT model, and our best alignment model (COMET + MT-HT). We observe that COMET + MT-HT scores are higher than the base MT model for all books, indicating that our alignment method makes the translations more lexically diverse. It is interesting to see that our method brings the results closer to or even exceeds human translation in terms of lexical diversity on some books (e.g. 5, 9, 14, and 16). Overall, the MTLD scores of the alignment models are between those of the base MT model and human translation.

\subsection{Impact of Hyper-parameter}
To examine the impact of hyper-parameter $\beta$ (see Section~\ref{sec:mtreduction}), we report the results when it is set to 0.0, 
i.e.\ only considering the reward learning. 
Models
trained without the constraint of negative log-likelihood loss lead, as expected, to worse content scores across the board as they move too far from the base MT model (Table \ref{tab:hyper}). These models achieve better classification scores but worse naturalness results (except MTLD, B1, and CDU in MT-OR and MTLD in MT-HT). The higher scores on classifiers could be due to characteristics of translated language beyond those related to high lexical diversity. Future work is needed to determine how 
the classifiers, lexical diversity, machine translationese and naturalness are precisely related.

\section{Conclusion}
We proposed a reinforcement learning based alignment framework for machine translation, which improves translation quality from multiple perspectives. Using the evaluation model COMET and different binary translationese classifiers trained with MT, HT, and original target-language data as reward models, we approximate human preference and align the MT model with it. Our experiments on English-to-Dutch literary translation show that our model produces translations that are lexically richer and more natural without loss in translation accuracy.

\section*{Limitations}
Due to the computational resources required
, we were only able to perform extensive experiments on one language pair and domain. Since we first wanted to show that our method is sound in a simple setting, i.e.\ training a model from scratch, we have not proceeded to involve complex settings and computationally-heavy models, such as pre-trained large language models. 
Furthermore, our metrics for evaluating naturalness are mostly limited to lexical diversity, while writing style in general is much broader and difficult to capture with automatic metrics. We acknowledge that large-scale human evaluation, beyond our surface-level inspection in Section \ref{sec:surface}, could bring important insights.

\section*{Acknowledgments}

The anonymous reviewers of ACL Rolling Review provided us with useful comments which contributed to improving this paper and its presentation, so we’re grateful to them. We would also like to thank the Center for Information Technology of the University of Groningen for their support and for providing access to the high-performance computing cluster Hábrók. Huiyuan Lai is supported by the Sector Plan in the Humanities ``Humane AI'' funding from the Dutch Ministry.
Antonio Toral is supported by a Beatriz Galindo senior fellowship (BG23/00152) from the Spanish Ministry of Science and Innovation.

\bibliography{custom}

\onecolumn

\appendix

\section{Appendix}
\label{sec:appendix}

\subsection{Test Set Novels}
\label{app:test}

\begin{table}[h!]
\begin{center}
\scalebox{0.78}{
\begin{tabular}{rllrl}
\toprule
\bf ID &\bf Author &  \bf Title & \bf Year Published & \bf Genre \\
\midrule
1&Patricia Highsmith & Ripley Under Water & 1991 & Thriller, suspense \\ 
2&J.D. Salinger & The Catcher in the Rye & 1951 & Literary fiction \\ 
3&Mark Twain & Adventures of Huckleberry Finn & 1884 & Literary fiction \\ 
4&John Steinbeck & The Grapes of Wrath & 1939 & Literary fiction \\ 
5&John Boyne & The Boy in the Striped Pyjamas & 2006 & Historical fiction \\ 
6&Nicci French & Blue Monday: A Frieda Klein Mystery & 2011 & Thriller, suspense \\ 
7&Philip Roth & The Plot Against America & 2004 & Political fiction \\ 
8&Paul Auster & Sunset Park  & 2010 & Literary fiction \\
9&Khaled Hosseini & A Thousand Splendid Suns & 2007 & Literary fiction \\ 
10&George Orwell & 1984 & 1949 & Literary fiction\\ 
11&John Irving & Last Night in Twisted River & 2009 & Literary fiction\\ 
12&E.L. James & Fifty Shades of Grey & 2011 & Erotic thriller \\ 
13&Jonathan Franzen & The Corrections & 2001 & Literary fiction \\ 
14&Stephen King & 11/22/63 & 2011 & Science-fiction \\ 
15&Oscar Wilde & The Picture of Dorian Gray & 1890 & Literary fiction \\ 
16&John Grisham & The Confession & 2010 & Thriller, suspense \\ 
17&William Golding & Lord of the Flies & 1954 & Literary fiction \\ 
18&Irvin D. Yalom & The Spinoza Problem & 2012 & Historical fiction \\ 
19&J.R.R Tolkien & The Return of the King & 1955 & Fantasy \\ 
20&David Baldacci & Divine Justice & 2008 & Thriller, suspense\\
21&Julian Barnes & The Sense of an Ending & 2011 & Literary fiction \\ 
22&James Patterson & The Quickie & 2007 & Thriller, suspense \\ 
23&Sophie Kinsella & Shopaholic and Baby & 2007 & Popular literature \\ 
24&J.K. Rowling & Harry Potter and the Deathly Hallows & 2007 & Fantasy \\ 
25&John le Carré & Our Kind of Traitor & 2010 & Thriller, spy fiction \\ 
26&Jack Kerouac & On the Road & 1957 & Literary fiction \\ 
27&Karin Slaughter & Fractured & 2008 & Thriller, suspense \\ 
28&Ernest Hemingway & The Old Man and the Sea & 1952 & Literary fiction \\ 
29&David Mitchell & The Thousand Autumns of Jacob de Zoet & 2010 & Historical fiction \\ 
30&James Joyce & Ulysses & 1922 & Literary fiction \\ 
31&Thomas Pynchon & Gravity's Rainbow & 1973 & Historical fiction\\ 
\bottomrule
\end{tabular}
}
\caption{Information on test set books.}
\label{table:test_set_info}
\end{center}
\end{table}

\subsection{Post-processing Examples}
\label{app:post}
\begin{table}[h!]
\begin{center}
\scalebox{0.9}{
\begin{tabular}{p{8.5cm}|p{8.5cm}}
\toprule
\bf Original Outputs & \bf Post-processed outputs\\
\hline
Bijna een jaar lang heeft hij foto's genomen van verlaten dingen............. & Bijna een jaar lang heeft hij foto's genomen van verlaten dingen.\\
Ongetwijfeld mag hij blij zijn dat hij deze baan heeft gevonden........ & Ongetwijfeld mag hij blij zijn dat hij deze baan heeft gevonden.\\
In het begin was hij verbijsterd door de wanorde en de vuiligheid, de verwaarlozing.............. & In het begin was hij verbijsterd door de wanorde en de vuiligheid, de verwaarlozing.\\
\bottomrule
\end{tabular}
}
\caption{Post-processing examples.}
\label{table:post-processing}
\end{center}
\end{table}

\end{document}